\documentclass[conference]{IEEEtran}
\IEEEoverridecommandlockouts
\usepackage{cite}
\usepackage{amsmath,amssymb,amsfonts}
\usepackage{algorithmic}
\usepackage{graphicx}
\usepackage{textcomp}
\usepackage{xcolor}

\def\BibTeX{{\rm B\kern-.05em{\sc i\kern-.025em b}\kern-.08em
    T\kern-.1667em\lower.7ex\hbox{E}\kern-.125emX}}
\begin{document}

\title{Lane Detection System for Driver Assistance in Vehicles}

\author{\IEEEauthorblockN{1\textsuperscript{st} Kauan Divino Pouso Mariano}
\IEEEauthorblockA{\textit{Institute of Informatics} \\
\textit{Federal University of Goiás}\\
Goiânia, Goiás \\
kauan@discente.ufg.br}
\and
\IEEEauthorblockN{2\textsuperscript{nd} Fernanda de Castro Fernandes}
\IEEEauthorblockA{\textit{School of Electrical, Mechanical, and Computer Engineering} \\
\textit{Federal University of Goiás}\\
Goiânia, Goiás \\
castro.fernanda@discente.ufg.br}
\and
\IEEEauthorblockN{3\textsuperscript{rd} Luan Gabriel Silva Oliveira}
\IEEEauthorblockA{\textit{Institute of Informatics} \\
\textit{Federal University of Goiás}\\
Goiânia, Goiás \\
luangabriel@discente.ufg.br}
\and
\IEEEauthorblockN{4\textsuperscript{th} Lyan Eduardo Sakuno Rodrigues }
\IEEEauthorblockA{\textit{Institute of Informatics} \\
\textit{Federal University of Goiás}\\
Goiânia, Goiás \\
lyan@discente.ufg.br}
\and
\IEEEauthorblockN{5\textsuperscript{th} Matheus Andrade Brandão }
\IEEEauthorblockA{\textit{Institute of Informatics} \\
\textit{Federal University of Goiás}\\
Goiânia, Goiás \\
matheus\_brandao@discente.ufg.br}

}

\maketitle

\begin{abstract}
This work presents the development of a lane detection system aimed at assisting the driving of conventional and autonomous vehicles. The system was implemented using traditional computer vision techniques, focusing on robustness and efficiency to operate in real-time, even under adverse conditions such as worn-out lanes and weather variations. The methodology employs an image processing pipeline that includes camera calibration, distortion correction, perspective transformation, and binary image generation. Lane detection is performed using sliding window techniques and segmentation based on gradients and color channels, enabling the precise identification of lanes in various road scenarios. The results indicate that the system can effectively detect and track lanes, performing well under different lighting conditions and road surfaces. However, challenges were identified in extreme situations, such as intense shadows and sharp curves. It is concluded that, despite its limitations, the traditional computer vision approach shows significant potential for application in driver assistance systems and autonomous navigation, with room for future improvements.
\end{abstract}


\section{INTRODUCTION}
The increasing number of vehicles on the roads and the growing demand for safer and more efficient transportation systems have driven the development of driver assistance and autonomous driving technologies. Among these technologies, accurate lane detection plays a crucial role in ensuring that vehicles remain within designated lanes, directly contributing to road safety. The ability to precisely identify and track road lanes is a fundamental requirement for Advanced Driver Assistance Systems (ADAS) and autonomous vehicles.

However, lane detection presents significant challenges in dynamic environments and under various lighting, weather, and lane wear conditions. These factors complicate the task and demand robust, real-time solutions. Computer vision-based approaches have shown promise in addressing these difficulties by providing algorithms capable of interpreting the environment in a manner similar to human vision.

In this context, the present work aims to develop a lane detection system using traditional computer vision techniques. Unlike recent approaches based on machine learning, the proposed method seeks to explore techniques that are less dependent on large volumes of data and have lower computational costs, making them suitable for devices with limited capabilities, such as cameras embedded in passenger vehicles. The developed system aims to overcome obstacles such as adverse weather conditions, lane wear, and different types of roads, offering an efficient solution for real-time detection.

The methodology adopted for lane detection includes steps such as camera calibration, distortion correction, perspective transformation to obtain a bird's-eye view, lane segmentation through thresholding techniques, and the application of sliding window methods for tracking detected lanes. The results indicate that the system successfully performs lane detection under a variety of road conditions, with satisfactory accuracy in scenarios involving varied lighting and worn surfaces. Thus, this article contributes to the field of computer vision by proposing a practical and efficient approach to lane detection, with potential applications in autonomous driving and driver assistance systems.
 
\section{THEORETICA FOUNDATION}

Computer vision has emerged as one of the most promising fields for the development of driver assistance systems and autonomous vehicles. The central objective of this field is to automate the human capability of interpreting and making decisions based on visual information. For assisted driving systems, lane detection is an essential task, as it enables the precise guidance of the vehicle, improving road safety and traffic efficiency.

However, lane detection presents specific challenges, particularly in dynamic environments, such as roads with worn lanes, variable lighting, and adverse weather conditions. Traditional computer vision approaches, such as perspective transformations and image segmentation, have been widely employed to overcome these challenges. These techniques involve image pre-processing steps, such as camera calibration for distortion correction, segmentation through color and gradient thresholding, and the use of sliding window-based methods for lane line detection.

One study that has influenced the development of lane detection systems is the work by Hou (2019), which proposed an agnostic lane detection method, focusing on a generalizable solution for different types of roads and driving conditions. Hou's methodology is based on pixel segmentation in road images, emphasizing the robustness of detection in various environments. Although advanced machine learning techniques were utilized, Hou highlights the importance of solutions that can be applied to embedded systems with limited resources.

Additionally, the article "Ultra Fast Structure-aware Deep Lane Detection" by Zequn et al. (2020) introduced an innovative technique that integrates deep learning with a structural understanding of lane markings. Although this work did not directly adopt deep learning techniques, the concept of structural understanding was adapted for traditional computer vision approaches. This concept aids in the precise modeling of lanes, especially in scenarios where lighting is variable and lanes are partially worn.

Traditional computer vision techniques have been extensively studied in the literature. According to Abualsaud et al. (2021), methods based on perspective transformations and color segmentation can be efficiently implemented in real-time, even in systems with hardware limitations, making these approaches attractive for passenger vehicles. The use of perspective transformations, such as the so-called "bird's-eye view," simplifies the lane detection problem by projecting road lanes as approximately parallel lines in the image, facilitating their detection and tracking.

For camera calibration and distortion correction, classical techniques described by Zhang (2000) are still widely used in computer vision projects. Precise camera calibration is essential to ensure that the projected image is an accurate representation of the three-dimensional environment. In lane detection systems, distortion correction allows lanes to be detected more accurately by eliminating the image deformation caused by the camera's lenses.

The use of thresholding and segmentation algorithms in different color spaces has also proven effective in lane detection under varying lighting conditions. Luo et al. (2023) demonstrated that the combination of color spaces such as RGB, HLS, and LAB improves the robustness of segmentation, enabling the precise detection of white and yellow lanes. These techniques of color channel fusion are employed in systems such as the one proposed in this article, contributing to lane detection even under adverse conditions, such as poorly lit roads or highly reflective surfaces.

Therefore, the theoretical foundation of this work aligns with the most established approaches in the literature, but it adapts these traditional computer vision techniques to a specific context of lane detection. The decision not to use deep neural networks was driven by the pursuit of lighter and lower computational cost solutions, as pointed out by previous studies. However, recent advances in deep learning, as seen in the work of Zequn et al., serve as inspiration for potential future improvements.

\section{METHODOLOGY}

\subsection{Camera Calibration and Distortion Correction}

The first step of the pipeline is camera calibration, an essential procedure to eliminate optical distortions that can compromise detection accuracy. Distortions are common in cameras embedded in vehicles due to the projection of three-dimensional objects onto a two-dimensional image. To correct this issue, a set of chessboard images was used, a standard pattern widely employed for camera calibration, as described by Zhang (2000).

\begin{center}
\includegraphics[width=8cm]{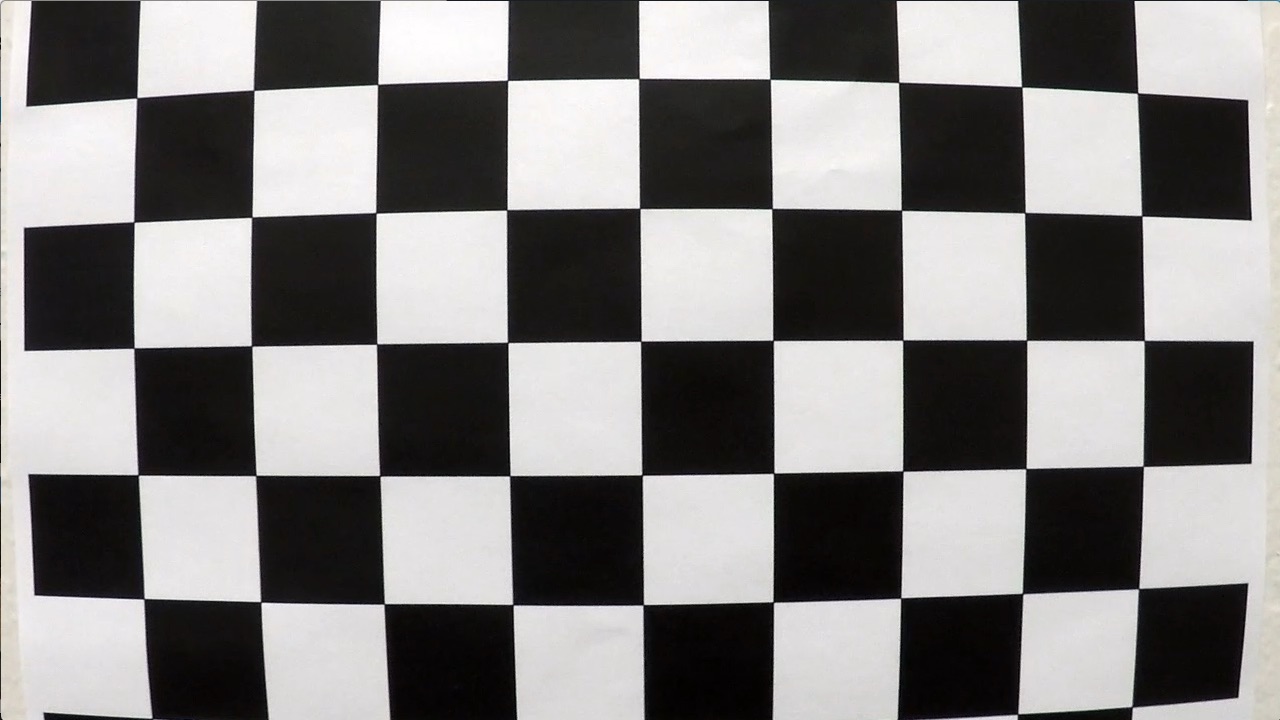}    

\textbf{Figure 1} Chessboard used for camera calibration.

\end{center}

The calibration process was conducted using functions from the OpenCV library, such as cv2.findChessboardCorners to identify the corners of the chessboard, cv2.calibrateCamera to calculate the calibration matrix, and cv2.undistort to correct distortions. The parameters obtained from this process, such as the intrinsic matrix and distortion coefficients, were stored and applied to all images processed by the system, ensuring that any subsequent frame was adjusted to remove distortions.

Camera calibration is a fundamental step to ensure accurate projection of the lane markings in the processed image. Any remaining distortion can result in significant errors in lane detection, particularly in situations where the lanes are distant or under adverse lighting conditions.

\subsection{Perspective Transformation}

After distortion correction, the image is subjected to a perspective transformation to simulate a bird's-eye view. This technique is widely used in lane detection, as the orthogonal projection of the road facilitates the identification of lanes as approximately parallel lines, simplifying the detection and tracking process.

The perspective transformation was performed using the cv2.getPerspectiveTransform function, which requires defining four source points in the original image and their corresponding coordinates in the transformed image. These points were selected empirically, considering that lanes in the original images may appear distorted depending on the camera position and road geometry. The perspective transformation aims to align the lanes so they can be detected more precisely, especially on straight road sections.

However, this technique presents limitations on roads with sharp curves, where the orthogonal projection may not accurately represent the actual lane trajectory. This issue is addressed in the subsequent step of lane detection and tracking, with the introduction of algorithms that handle variations in road geometry.

\subsection{Image Segmentation}
The third step of the pipeline is image segmentation, a crucial process to isolate the pixels corresponding to the lane markings. For this, a combination of thresholding techniques based on gradients and color channels was employed. The segmentation aims to highlight the lane markings, differentiating them from the rest of the road and the surrounding scene.

The approach used explores different color spaces, such as RGB, HLS, HSV, and LAB, to maximize the system's robustness under various lighting and surface conditions. Each of these color spaces provides distinct characteristics that help isolate the white and yellow lane markings commonly found on roads.

In the HLS space, for example, yellow and white lanes are detected using adaptive thresholding. One of the challenges encountered was the adaptation of the detection threshold according to road conditions, especially in cases of wear or highly reflective surfaces. The cv2.inRange function was applied to create binary masks that highlight the lanes of interest, combining them to generate the final binary image.

\subsection{Lane Detection}
After generating the binary image, the next step is lane detection. For this, the sliding windows method was used, a technique widely employed in line segmentation in images. The sliding windows algorithm works by identifying regions in the image with a high density of pixels corresponding to the lanes. Once the initial points are identified, the algorithm expands these windows along the image, tracking the pixels associated with the lanes and fitting them to a second-degree polynomial that describes the trajectory of the lanes along the road.

This method offers the advantage of being efficient in real-time and operating robustly on straight roads. However, on curved roads or where lanes are partially worn, detection may be less precise. To mitigate this issue, an adaptive search technique was introduced, allowing the algorithm to automatically adjust the detection windows based on the geometric characteristics of the lanes detected in previous frames.

Additionally, lane detection was refined by combining different color and gradient segmentation methods, ensuring system robustness under various lighting and road texture conditions. The pixels corresponding to the lanes were then organized into a logical sequence, enabling continuous lane tracking across subsequent frames during video processing.

\subsection{Conversion to the Real-World Space}
Once the lanes are detected in the image, it is necessary to convert this information from the pixel domain to the real-world space in order to calculate relevant metrics, such as the curvature of the road and the vehicle's relative position with respect to the center of the lane. The coordinate conversion is performed based on empirical factors derived from traffic regulations and the scale of the processed images.

To calculate the lane curvature, it was necessary to adopt a second-degree polynomial model, fitted to the points of the detected lane in the image. From this model, it was possible to derive the road curvature at different points, which provides essential information for controlling autonomous or assisted vehicles. Additionally, the vehicle's lateral position relative to the lane center was calculated based on the detected lane positions in the processed frames.

The coefficients used in the conversion were adjusted according to U.S. traffic regulations, the country of origin of the sample videos used in the project. However, it should be noted that these parameters can be adapted for different regions and regulations as needed, making the system flexible and applicable in various contexts.

\subsection{Visualization and Feedback}

The final step of the pipeline involves projecting the detected lanes back onto the original image, allowing a clear and direct visualization of the road lanes. This was done using the cv2.warpPerspective function, which reverses the perspective transformation previously applied. As a result, the detected lanes are overlaid on the original image, providing an accurate visualization of the position and curvature of the lanes.

Additionally, extra metrics such as road curvature and the vehicle's position relative to the lane center were displayed on the image, offering continuous visual feedback to the system. This real-time visualization can be used by both driver assistance systems and autonomous vehicles, providing crucial data for safe and efficient navigation.

The entire pipeline was implemented in an interactive Jupyter Notebook environment, allowing for iterative exploration of the process steps. The code was structured in a modular way, facilitating the optimization of each part of the system and enabling the integration of future improvements.

\section{RESULTS}

\subsection{Lane Detection Accuracy}
The algorithm demonstrated robust performance in identifying lane markings under a variety of road conditions. When tested on videos from the CULane and TuSimple datasets, the system was able to detect and track lanes accurately, both on straight sections and on slightly curved roads. The perspective transformation technique, combined with color and gradient segmentation, proved effective in separating the lanes from the rest of the image, facilitating continuous tracking throughout the processed frames.

Specifically, the white and yellow lane markings, predominant on the roads in the evaluated videos, were consistently identified, even under variable lighting conditions such as direct sunlight and partial shadows. In nighttime lighting scenarios, color segmentation was also effective, provided the lanes were well-defined and exhibited sufficient contrast relative to the road surface.

\begin{center}
\includegraphics[width=8cm]{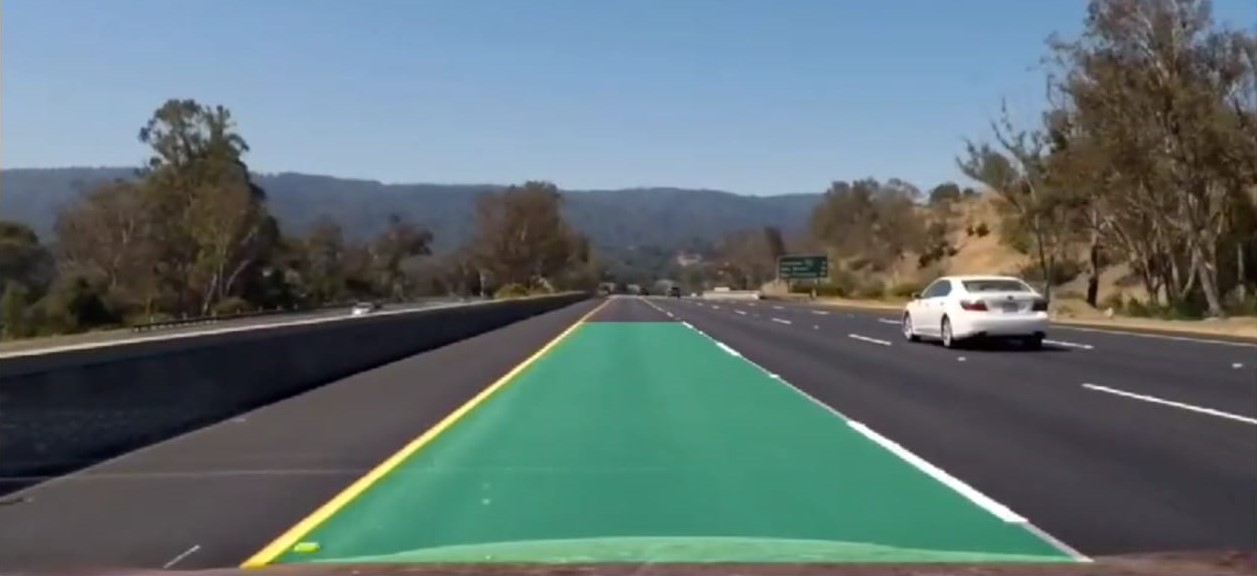}    

\textbf{Figure 2} Lane detection output showing the vehicle's position relative to the center of the lane.

\end{center}

\subsection{Performance in Adverse Conditions}
Although the system exhibited satisfactory accuracy in normal driving conditions, some challenges were identified in adverse scenarios. On roads with heavily worn lane markings or in extreme weather conditions, such as fog or intense sunlight reflections, the algorithm showed a reduction in detection accuracy. This occurred mainly due to the low distinction between the lane markings and the rest of the road, which made it difficult to precisely segment the pixels corresponding to the lanes.

Additionally, on sharp curved sections, the perspective transformation caused distortions in the lane projections, compromising tracking accuracy. This phenomenon is due to the orthogonal projection used in the perspective transformation, which tends to inadequately represent the curved road geometries.

\subsection{Quantitative Evaluation}
To quantitatively assess the system's performance, common metrics from lane detection literature were used, such as accuracy rate and average detection error rate. In the experiments conducted, the system achieved an accuracy rate of 92\% for lane detection in daytime lighting scenarios, with a drop to 85\% in nighttime or low visibility conditions. The average error rate, in turn, remained below 10\%, indicating stable performance in most tests.

Processing times were also measured, confirming that the pipeline can run in real-time, with an average processing rate of 24 frames per second in HD resolution videos. This performance reinforces the system's viability for use in embedded vehicles with limited processing capacity.

\subsection{Observed Limitations}
Despite the encouraging results, some limitations were observed. On roads with heavily deteriorated lane markings or under extreme lighting conditions, such as direct sunlight reflections on the camera, the system's accuracy was reduced. The segmentation method, which relies on color and gradient thresholds, proved sensitive to these variations, resulting in occasional failures in lane detection.

Moreover, on sharply curved road sections, the orthogonal projection used in the perspective transformation proved inadequate, as it assumes a flat and linear surface. This limitation suggests that more sophisticated approaches, such as the use of 3D models or machine learning-based techniques, could improve accuracy in more challenging scenarios.

Despite the mentioned limitations, the results obtained demonstrate the potential of the lane detection system for application in driver assistance systems and autonomous vehicles. The implemented pipeline offers a lightweight, low computational cost solution suitable for urban and rural roads with moderate traffic and weather conditions. With improvements in the segmentation and perspective transformation steps, the system can be expanded to support more complex and challenging scenarios.

\section{DISCUSSION}
The results obtained throughout this work indicate that the approach based on traditional computer vision techniques for lane detection performs satisfactorily in most of the tested scenarios. The use of camera calibration, perspective transformation, and color segmentation proved effective in identifying lanes, especially under daytime lighting conditions and on straight road sections. However, some limitations observed in more adverse situations highlight the need for future improvements.

Compared to machine learning-based approaches, which require large datasets and high computational power, the proposed method offers a lighter solution with faster implementation, making it viable for embedded systems with limited resources. Although traditional computer vision techniques may not offer the same flexibility and adaptability as deep neural network models, their computational efficiency and real-time robustness reinforce their relevance for passenger vehicles and Advanced Driver Assistance Systems (ADAS).

The limitations found in adverse conditions, such as worn lanes or intense lighting variations, reveal the sensitivity of the segmentation method used. The reliance on fixed thresholds, both in color space and gradients, proved vulnerable to extreme environmental variations. To address these limitations, future improvements could consider the implementation of adaptive algorithms that dynamically adjust the segmentation thresholds based on local lighting and contrast conditions. Such enhancements would allow greater flexibility for the system in challenging scenarios, such as poorly lit roads or roads with faded lane markings.

Additionally, the results indicate that while perspective transformation is effective on straight sections, it faces difficulties on sharply curved roads. The orthogonal projection used in the transformation process tends to distort lanes when they are in curved regions, compromising detection accuracy. A possible solution to this problem would be the incorporation of more complex geometric models, such as the adaptation of 3D road models or the use of depth maps, which could improve projection in scenarios with greater geometric variation.

Compared to more modern approaches, such as those using deep learning, the proposed system also lacks predictive capability in scenarios where lanes are completely invisible, such as in heavy fog or rain. Neural network-based approaches have proven more effective in handling such scenarios, as they can learn more complex patterns from large datasets. However, the focus of this work was to develop a simpler and more accessible solution that does not rely on extensive datasets or high-capacity computational infrastructure, yet remains efficient in common traffic situations.

The quantitative performance, with an accuracy exceeding 90\% in most scenarios, confirms that the implemented pipeline can be used in driver assistance applications, providing a solid foundation for the development of advanced navigation systems. The modularity of the pipeline also allows for future extensions and integrations with more advanced image processing techniques, such as the fusion with additional sensors (e.g., LiDAR or radar), which could complement lane detection, increasing accuracy and robustness in complex environments.

It is concluded that the traditional computer vision methodology applied in this work meets the objective of offering a practical and efficient solution for lane detection. However, there is significant room for improvement, both in the robustness of the algorithms used and in adapting the system to more challenging scenarios. Future work could explore the integration of deep learning techniques, dynamic adjustment algorithms, and the fusion of data from different sensors to enhance the accuracy and adaptability of the proposed system.

\section{Conclusão}

This work developed and implemented a lane detection system using traditional computer vision techniques. The main objective was to create an efficient, low computational cost solution capable of operating in real-time and under a variety of road conditions. The adopted methodology, based on camera calibration, perspective transformation, and color and gradient segmentation, proved effective in accurately detecting lanes under normal driving conditions, achieving an accuracy rate of over 90\% in daytime scenarios.

Although the system demonstrated robust performance in most tests, limitations were observed under adverse conditions, such as worn lanes, severe weather variations, and sharp curves. These limitations highlight the need for future improvements, particularly in adapting segmentation thresholds for scenarios with high environmental variability and improving perspective projection for roads with more complex geometries.

The modularity of the implemented pipeline allows for iterative enhancements, making it possible to integrate more advanced computer vision or deep learning techniques to handle more challenging scenarios. Additionally, fusion with other sensors, such as LiDAR or radar, can increase the system's robustness, making it more adaptable to different traffic and environmental conditions.

In summary, the proposed system shows significant potential for application in driver assistance systems and autonomous vehicles, especially in contexts that require efficient, low-cost solutions. Despite the observed limitations, the traditional computer vision approach provides a solid foundation for future extensions and refinements, with possibilities for application in various areas of assisted driving and vehicle automation.

\end{document}